\documentclass[letterpaper]{article} % DO NOT CHANGE THIS
\usepackage{aaai23}  % DO NOT CHANGE THIS
\usepackage{times}  % DO NOT CHANGE THIS
\usepackage{helvet}  % DO NOT CHANGE THIS
\usepackage{courier}  % DO NOT CHANGE THIS
\usepackage[hyphens]{url}  % DO NOT CHANGE THIS
\usepackage{graphicx} % DO NOT CHANGE THIS
\usepackage{natbib}  % DO NOT CHANGE THIS AND DO NOT ADD ANY OPTIONS TO IT
\usepackage{caption} % DO NOT CHANGE THIS AND DO NOT ADD ANY OPTIONS TO IT
\usepackage{algorithm}
\usepackage{algorithmic}
\usepackage{amsfonts}

\usepackage{newfloat}
\usepackage{listings}
\DeclareCaptionStyle{ruled}{labelfont=normalfont,labelsep=colon,strut=off} % DO NOT CHANGE THIS
\floatstyle{ruled}
\newfloat{listing}{tb}{lst}{}
\floatname{listing}{Listing}
\usepackage{tikz}
\usepackage{booktabs}
\usepackage{amsmath}
\usepackage{csquotes}

\title{Improved Earth Observation Satellite Scheduling \\ with Graph Neural Networks and Monte Carlo Tree Search}
\author{
    Antoine Jacquet\textsuperscript{\rm 1},
    Guillaume Infantes\textsuperscript{\rm 1},
    Emmanuel Benazera\textsuperscript{\rm 1},\\
    Vincent Baudoui\textsuperscript{\rm 2},
    Jonathan Guerra\textsuperscript{\rm 2},\\
    Stéphanie Roussel\textsuperscript{\rm 3}
}
\affiliations{
    \textsuperscript{\rm 1} Jolibrain, Toulouse, France \{firstname.lastname\}@jolibrain.com\\
    \textsuperscript{\rm 2} Airbus Defence \& Space, Toulouse, France \{firstname.lastname\}@airbus.com\\
    \textsuperscript{\rm 3} DTIS, ONERA, Université de Toulouse, France \{firstname.lastname\}@onera.fr \\
}

\usepackage{bibentry}
\begin{document}

\maketitle

\begin{abstract}

   Earth Observation Satellite Planning (EOSP) is a difficult optimization problem with considerable practical interest.  A set of requested observations must be scheduled on an agile Earth observation satellite while respecting constraints on their visibility window, as well as  maneuver constraints that impose varying delays between successive observations.  In addition, the problem is largely oversubscribed: there are much more candidate observations than can possibly be achieved.  Therefore, one must select the set of observations that will be performed while maximizing their cumulative benefit and propose a feasible schedule for these observations.  As previous work mostly focused on heuristic and iterative search algorithms, %we use a promising Neural Combinatorial Optimization approach to tackle this problem.  This
  this paper presents a new technique for selecting and scheduling observations based on Graph Neural Networks (GNNs) and Deep Reinforcement Learning (DRL).   GNNs are used to extract relevant information from the graphs representing instances of the EOSP, and DRL drives the search for optimal schedules.  A post-learning search step based on Monte Carlo Tree Search (MCTS) is added that is able to find even better solutions.
   Experiments show that it is able to learn on small problem instances and generalize to larger real-world instances, with very competitive performance compared to traditional approaches.% currently being deployed on real satellites.
\end{abstract}

\section{Introduction}

An Earth observation satellite (EOS) must acquire photographs of various locations on the surface of Earth to satisfy user requests.  An \emph{agile} EOS has degrees of freedom allowing it to target locations that are not exactly at its vertical in an earth-bound referential (\textquote{nadir})%(the point directly under the satellite is called its "nadir" point)
.  The satellite we consider is in low orbit; as a consequence, each observation is available in a visibility time window (VTW) that is significantly larger than its acquisition duration.  Maneuvering the satellite between two observations consists of modifying its pitch, yaw and roll angles triple---also called its \emph{attitude}---and thus implies delays that depend on the start and end observation targets as well as on the maneuver start date \cite{time_dependent_eosp}.  In addition, an agile EOS is typically oversubscribed: there are more observations to be performed that can possibly be achieved in the given operation temporal horizon.  As different acquisitions may be associated with different priorities or utilities, the Earth observation satellite planning problem (EOSP) consists in selecting a set of acquisitions that maximize their weighted cumulative values and designing a schedule for acquiring these observations while respecting the operational constraints.

The most complex instances of the EOSP involve several satellites orbiting the Earth over multiple orbits, and dependencies between targets \cite{EOSP_20_years}. For instance, an acquisition may consist of several pictures of the same earth-bound location to be taken by different satellites, and/or in different time-windows \cite{squillaci2023scheduling}.  In this paper, we limit our study to single-satellite, single-orbit, and single-shot problems, where there is only one satellite to control over a single orbit, and each acquisition is made of a single picture to be taken in its given VTW. Nevertheless, the problem is NP-complete \cite{lemaitre2002selecting}, and there is no practical solution to compute optimal schedules for problems of realistic size which contain a few thousand candidate acquisitions, thus focusing previous work towards approximate, heuristic and random search algorithms \cite{EOSP_20_years}.  Variants of the greedy randomized adaptive search procedure (GRASP) \cite{grasp} are commonly  deployed in practical applications. In \cite{Pralet23}, the author considers a single satellite scheduling problem with time-dependent maneuvers and aims at minimizing the tardiness associated with a set of observations that must be performed. The solving approach combines Dynamic Programming and Large Neighborhood Search techniques. The problem we consider here is different as decisions must not only be made on the observations scheduling order but also on their presence in the final scheduling. 

At the same time, the field of combinatorial optimization is currently the subject of an accrued interest from researchers in deep learning.  In particular, Deep Reinforcement Learning (DRL) offers a framework for learning heuristics for NP-complete problems that has been successfully applied to a wide range of problems \cite{nco}.  Following this trend, we build on previous work using a state-of-the-art combination of graph neural networks (GNN) and DRL.  This approach code-named \emph{Wheatley} was developed to address Job Shop Scheduling Problems with duration uncertainty \cite{wheatley-jssp}.  Here we adapt it to the EOSP and prove its efficiency in solving deterministic but largely over-subscribed problems.  Our simulation results show that we outperform the currently  deployed techniques.

Our main contributions are as follows:
(1) propose a graph search representation of the problem without time discretization;
(2) use deep reinforcement learning to solve the problem;
(3) use directly the problem graph representation as observations within the graph neural network;
(4) develop a post-training search phase based on MCTS that substantially improves the results.
The main outcomes are:
(1) very competitive results compared to baselines;
(2) good generalization abilities allowing to train on small instances and solve efficiently large instances.

The paper is organized as follows. First, we give a quick survey of related work based on deep learning approaches to solve the EOSP. Then, in Section \ref{sec:problem_representation}, we introduce the problem and discuss various representations used in this work. Section \ref{sec:sol_technique} is dedicated to the description of the machine learning architecture used for optimization and post-training search.  We provide simulation results on large size real-world instances of the problem in Section \ref{sec:expes}. We finally conclude and discuss further research directions.

\section{Related Work}
\label{sec:related_works}

The EOSP has been subject to a large body of research, from communities as varied as aerospace and engineering, operational research, computer science, remote sensing and multidisciplinary sciences. %The role of this paper is not to survey all previous literature on the EOSP.  Instead,
We refer the reader to \cite{EOSP_20_years} for a survey of non-machine learning approaches to the problem, and we focus our attention on DRL based approaches to the EOSP.
Note that we address the time-dependent EOSP, where the duration of a transition between two observations varies with time.  This contrasts with most of previous literature that assumes constant, time-independent transition duration.

%\todo[inline]{Steph: il manque ici un commentaire sur le fait que vous prenez en compte la time-dependency. Je ne suis pas au fait des papiers en DRL mais côté RO, il y en a vraiment peu qui prennent cette spécificité en compte}

Peng et al.\ \cite{EOSP_LSTM} address a slightly different problem where observations are scheduled on board% and uploading it at the beginning of an orbit
.
A LSTM-based encoding network is used to extract features, and a classification network is used to make a decision. 
Dalin et al.\ \cite{Multiagent_DRL} solve multi-satellite instances by modeling them as a Multi-Agent Markov Decision Processes, then use a DRL actor-critic architecture.  The actor is decentralized, each satellite using a relatively shallow network to select its action.  The critic is centralized and implemented as a large recurrent network taking input from all satellites.  
Hermann et al.\ \cite{RL_Agile} also address the multi-satellite problem: a policy is trained in a single satellite environment on a fixed number of imaging targets,
and then deployed in a multi-satellite scenario where each spacecraft has its own list of imaging targets.  Local policies are learned using a combination of Monte Carlo Tree Search (MCTS) to produce trajectories, and supervised learning to learn Q-values using the trajectories produced by MCTS as training examples.
Finally, Chun et al.\ \cite{math11194059} present a very similar approach; the main difference is that the transition durations are approximated during the training phase, whereas in our approach they are precisely computed based on discrete date values before training. 

%\cite{framework_MEC, RL_Agile}

\section{Problem Representation}
\label{sec:problem_representation}

An instance of the EOSP is defined by a set of candidate observations $\mathcal{O}$, or acquisition requests, and a time-horizon $\tau$ (in this work, the duration of an orbit).  Each observation $i \in \mathcal{O}$ is associated with its fixed duration $d_i$ and its VTW $[e_i; l_i]$ such that $l_i \leq \tau$. 

The transition duration between two acquisitions $i$ and $j$ is a function $\Delta_{i,j}(t_i)$ of the starting time $t_i$ of the first observation \footnote{Strictly speaking, the function depends on the starting time of the maneuver, which is equal to $t_i + d_i$, where $d_i$ is a deterministic duration.}. A schedule $\sigma$ is a sequence of selected observations with associated starting time. It is represented by a single mapping that associates with each candidate observation $i \in \mathcal{O}$ its starting time $t_i^\sigma$, such that $t_i^\sigma = -1$ for all observations $i$ \emph{not} selected in schedule $\sigma$. When there is no ambiguity on the schedule $\sigma$ considered, we write $t_i$ instead of $t_i^\sigma$.
 %A schedule is a sequence of observations with associated starting time.   $\left( (i, t_i)  \right)_{i \in \{1, 2, \dots n\}}$. 
%\todo[inline]{Steph: pas hyper clair comme notation. Pas plus simple de voir un schedule comme une fonction $\sigma$ qui associe une date de démarrage $t_i$ à chaque acquisition $i$ ? On peut aussi donner une valeur par défaut ($-1$?) si l'acquisition n'est pas sélectionnée, ou alors définir une fonction booléenne qui renvoie vrai ou faux pour chaque acquisition.}
A schedule $\sigma$ is feasible if: 
 %\begin{itemize}
 	%\item 
 	(i) each scheduled observation starts and ends within its VTW: $\forall i \in \mathcal{O}$ such that $t_i \neq -1$, $t_i \in W_i$ where $W_i = [e_i; l_i - d_i]$; % and $t_i + d_i\in [e_i; l_i]$;
% \todo{Steph: il faut une convention pour la fin. En général, on a $t_i \geq e_i$ en $t_i + d_i < l_i$ et on peut écrire directement $t_i \in[e_i,l_i-d_i[$ (une seule équation au lieu de 2).  \\Nico: Pour la borne supérieure, il faut être consistent avec la definition de la VTW au-dessus (interval ferme).}
  %\item 
  (ii) the time gap between two successive selected observations is greater or equal to the transition delay: $\forall i,j \in \mathcal{O}^2$ such that $t_i \neq -1$, $t_j \neq -1$ and $t_i \leq t_j$, $t_{j} - t_i \geq d_i + \Delta_{i,j}(t_i)$.
  %\end{itemize}%$t_{i+1} - t_i \geq d_i + \Delta_{i,i+1}(t_i)$; 
%  \todo[inline]{Steph: Avant, on parle d'acquisitions $i$ $j$ comme si c'était les entrées du problème. Mais là, $i$ représente l'ordre dans la séquence calculée. Soit vous définissez explicitement l'ordre, soit vous modifiez l'équation en l'écrivant pour toute paire d'acquisitions sélectionnées dont la date de la 1ere est avant la date de la 2eme: $\forall i,j$ such that $t_i \neq -1$, $t_j \neq -1$ and $t_i \leq t_j$ then $t_{j} - t_i \geq d_i + \Delta_{i,j}(t_i)$}
 %  (iii) it terminates at or before the horizon: $t_n + d_n \leq \tau$. 
   
 %  \todo[inline]{Steph: même souci qu'avant avec $n$ qui ne représente pas l'acquisition n mais la n-ième de la séquence (ici la dernière - il faudrait d'ailleurs dire explicitement que $n$ acquisitions sont sélectionnées). Une manière simple de faire et d'ajouter une hypothèse qui est que $\forall i \in [1,n]$, $l_i \leq \tau$. }
   
	%\todo[inline]{Steph:Il me semble qu'il manque une hypothèse qui dit que la fonction de transition est bien faite, dans le sens où elle ne permet pas d'arriver plus tôt si on la déclenche plus tard. Même principe que pour le trafic et les bouchons, si on part au boulot à 9h au lieu de 8h30, on met peut-être moins de temps de trajet mais on arrive forcément plus tard que si on était parti à 8h30.  \\Nico: Je ne pense pas qu'on utilise cette hypoth\`ese (FIFO).  Dans le STN oui, mais l\`a non je pense.}
	
    Each observation is associated with a utility value $u_i$ in $\mathbb{R}^+$. The goal is then to find a feasible schedule that maximizes the cumulative utility of the observations it includes: $\mathit{maximize} \left( \sum_{i : t_i \neq -1} u_i \right)$.

%	\todo[inline]{Steph: j'aurais écrit cela de la manière en dessous, avec la notation  de sélection (je le mets en dehors du todo pour que ça compile)}  $\mathit{maximize} \sum_{\substack{i \in [1,n] \\ t_i \neq -1}} u_i$.

\subsection{Classical Approach : Time Discretization}
In the EOSP, the start time $t_i$  of each observation has a continuous domain ($W_i$), and the transition durations $\Delta_{i,j}(t_i)$ are continuous functions of continuous variables.  Therefore, the EOSP is not a pure discrete problem. However, it is often re-framed as such, either by making assumptions on the start time of transitions (for instance, every transition starts as soon as possible), or by crudely discretizing the time variable domains. 
 
%This work follows the later paradigm, where time is discretized using a fixed and constant time-step.
\paragraph{Discrete graph.} In this work, the problem input is provided under the form a very large time-discretized graph, later called the \textquote{discrete graph}, and denoted $\mathcal{G}^D$. More precisely, each visibility window $W_i$ associated with observation $i$ is discretized into a set of candidate starting dates denoted $W_i^D$. In graph $\mathcal{G}^D$, every candidate acquisition $i$ is represented $\mathit{card}(W_i^D)$ times. Formally, for each observation $i$ and each possible discretized starting time $t_i \in W_i^D$, $\mathcal{G}^D$ contains a node $(i, t_i)$ representing the fact that observation $i$ start date is equal to $t_i$. An arc between two nodes associated with observations $i$ and $j$ is present if the end observation $j$ is a possible immediate successor of the start observation $i$. The acquisition and transition durations are accounted for while defining the arcs: an arc between $(i, t_i)$ and $(j, t_j)$ is such that $t_j$ is the smallest discrete time in $W_j^D$ satisfying $t_i + d_i + \Delta_{ij}(t_i) \leq t_j$.  Note that there is an (implicit) mutual exclusion between two nodes $(i, t_i)$ and $(i, t_i')$, $t_i \neq t_i'$, to indicate that observation $i$ must not be performed twice. A virtual source node $(0,t_0)$ is added to represent the temporal horizon starting date. 

\begin{figure} [t]
\begin{center}
 \includegraphics[width=8cm]{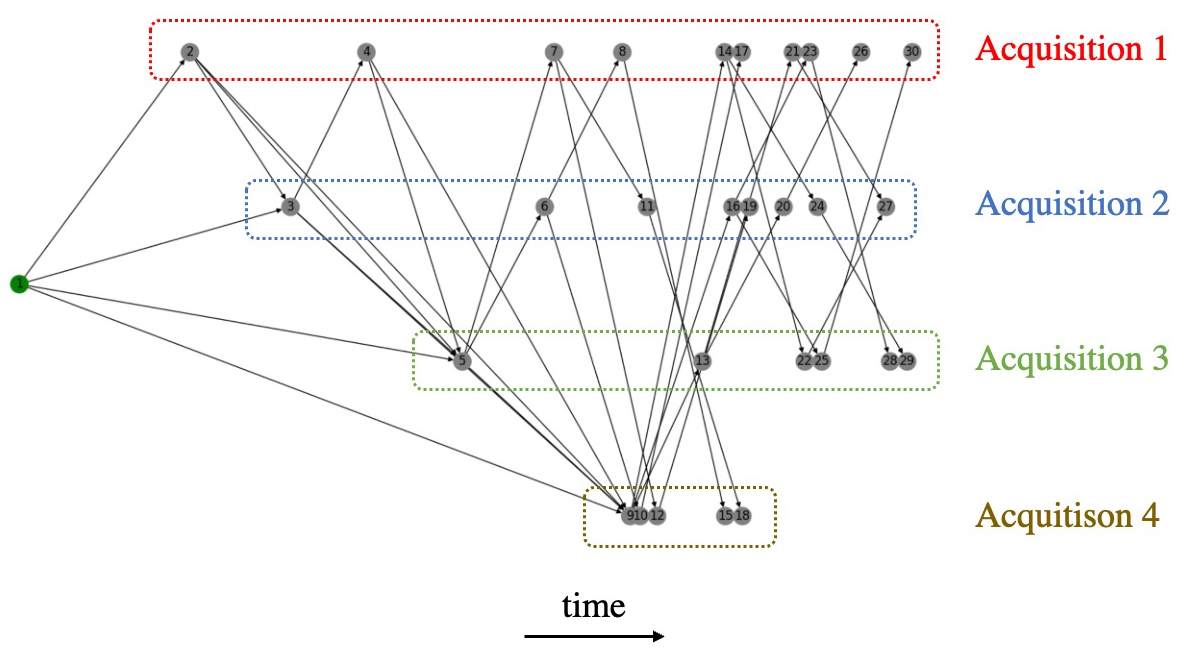}   
\end{center}
\caption{Discrete graph for 4 candidate acquisitions. Corresponding continuous graph is a 4-node clique.}
\label{fig:discrete_graph}
\end{figure}

Finally, some arcs are pruned using considerations of optimality: if an observation $k$ can be inserted between $(i, t_i)$ and $(j, t_j)$ without breaking the constraints of the problem (that is, the transition delays), then the arc between $(i, t_i)$ and $(j, t_j)$ is removed.  In this case, every path between the two nodes must go through one node $(k, t_k)$.  The reasoning is that every \emph{optimal} solution that includes both $(i, t_i)$ and $(j, t_j)$ would also include observation $k$.  Therefore, $(j, t_j)$ should not be an \emph{immediate} successor of $(i, t_i)$.  Figure \ref{fig:discrete_graph} provides an example of a discrete graph for a problem containing 4 candidate acquisitions.

 \paragraph{Schedule and graph update.} With such a discrete graph, a schedule can be represented as path starting from the source node. Formally, a schedule is a list of nodes $((0,t_0), (k,t_k), \ldots (n,t_n))$ in which $(0,t_0)$ is the source node and successive pairs of nodes in the path are arcs. Nodes that are not part of the path are considered as not selected.  
 %A schedule can be built by starting from a node representing the origin of the graph and following the edges to add observations in chronological order (the earliest are added to the schedule before the latest).  
 When building sequentially a schedule, the discrete graph can be simplified accordingly, as detailed as follows. Let $m$ denote the last node in a schedule $\sigma$. When adding a new node $n = (i, t_i)$ to $\sigma$, the graph is updated in the following way:
 (1) all arcs outgoing from node $m$ are removed, except the one leading to the newly scheduled node $n$; (2) all other nodes candidate for observation $i$ (nodes $(i,t'_i)$ with $t'i \neq t_i$) are deleted; (3) all nodes that are unreachable from $n$ are removed.
 
 The discrete graph is convenient for a typical state-space approach such as using an implicit enumeration algorithm (Dijkstra, A*) or Dynamic Programming. Our solution technique is based on representing the process of \emph{building an optimal schedule} for an instance of the EOSP as a reinforcement learning problem.%, then using DRL algorithms to solve it. 
 
 \subsection{Sequential Decision Model}
 \label{sec:mdp}
 
 Reinforcement learning is concerned with learning the solution of a \emph{Markov Decision Process} (MDP), which is a discrete-time sequential decision model.  An MDP is defined as a tuple $(S, A, T, R)$ where $S$ is the state space, $A$ the action space, $T$ the transition matrix, and $R$ the reward function \cite{puterman2014markov}. The definition of these elements flows directly from the discrete graph representation:
 \begin{itemize}
 	\item A state $s \in S$ is a discrete EOSP graph as defined before along with a schedule $\sigma$. It can be either the initial discrete graph for which the schedule is empty, or an updated graph along with a non-empty schedule; %intermediate graph where some nodes and edges have been selected to account for the fact that the beginning of the schedule has been decided;
 	\item Given a state $s$ and its associated schedule $\sigma$, the set of available actions $a$ is the set of all possible successors of the last node in $\sigma$;% (thus leading to \emph{chronological insertions}); 
 	\item MDPs naturally handle uncertainty in the problem. In the general case, it is represented in the transition matrix: $T(s, a, s') = \Pr(s(t+1)=s' \mid s(t)=s, a(t)=a)$.  However, our formulation of the EOSP is deterministic, therefore the MDP contains no uncertainty \footnote{Note that non-deterministic MDPs could be considered in the EOSP to account for the cloud cover uncertainty. Such an aspect is out of scope in this work.}.  Given an initial state $s$ (discrete graph with schedule) and a selected action $a$ (the next observation to add to the schedule), the transition matrix gives probability 1 to the state $s'$ representing the discrete graph after the addition of $a$ to the schedule, following the update procedure described previously. 	
 	% commentaire Jonathan 06/02/25: petit commentaire ici: si on a la place on pourrait dire que notre formulation du problème est déterministe mais que cet aspect des MDP pourrait être mis à profit pour prendre en compte les incertitudes météo. A voir si on a de la place et si ça vous va je peux rajouter une phrase sur ça.
 	\item As every inserted observation is feasible by definition, we use as an immediate reward the utility value associated with the selected observation i.e. $R(s,a,s') = u_a$.  The aim is to maximize the undiscounted sum of immediate rewards. 
 \end{itemize}
 
 These components define a fully observable MDP on which RL approaches can be based: the agent learns how to choose the next action in each state (policy) and the environment (or simulator) is responsible for providing a reward associated with each selected action and updating the state accordingly.
 Although convenient for state-space approaches, the discrete graph has the drawback of quickly becoming huge as the number of candidate acquisition grows, making it unsuitable as an input to a GNN-based agent. For this reason, we derive from the discrete graph $\mathcal{G}^D$ a (much) more compact graph $\mathcal{G}^C$ that we call the \textquote{continuous graph} and use it as an input for the agent. 
 
 %While , we do not provide a perfect description of the state $s$ to the deep neural network.  As explained above,  we do not feed the huge discrete graph to the neural network.  Instead, we use a continuous graph that is not a perfectly accurate representation of the problem,
 
\subsection{Our Approach : Continuous-Time Graph}  The continuous graph $\mathcal{G}^C$ is built upon the discrete graph $\mathcal{G}^D$ as follows. In $\mathcal{G}^C$, a node is simply defined by a candidate observation $i$, %each candidate observation is represented exactly once, 
with no mention of its exact starting time.  An arc $(i, j)$ is present in $\mathcal{G}^C$ if and only if there is an arc $((i, t_i), (j, t_j))$ in $\mathcal{G}^D$.  Note that this may lead to two-nodes-cycles if the corresponding observations may be performed in any order. In $\mathcal{G}^C$, the set of every cycle-free path is a super set of the possible schedules: %order of acquisitions (a super-set.
after selecting some acquisitions, VTW for later acquisitions may shrink to empty time windows, making these later acquisitions impossible to select.  

The continuous graph $\mathcal{G}^C$ cannot be used only by itself in the RL approach, as it does not contain the precise transition duration information. However, it is compact enough to be used as an input for a GNN-based agent. Therefore, our RL approach uses both graph representations. As illustrated on Figure~\ref{fig:archi}, the agent is fed with the compact continuous graph $\mathcal{G}^C$ and selects the next acquisition $i$ to be inserted in the schedule. To track which candidate observations are the possible next actions, the simulator uses the discrete graph $\mathcal{G}^D$ and assumes that observation $i$ is started at its earliest possible (discrete) time. The discrete graph $\mathcal{G}^D$ is updated with this decision as described previously, and these changes are then reflected in the continuous graph $\mathcal{G}^C$.  The resulting continuous graph is fed to the GNN-based agent at the next iteration.   %The main loop of our algorithm is presented in Alg.\ \ref{}.  
The set of candidate acquisitions that can be added to the current schedule is the set of immediate successors of the last node scheduled in the discrete graph $\mathcal{G}^D$. % The learning algorithm is responsible for selecting one particular acquisition among them, using the procedure detailed later.
Note that node features of the continuous graph contain information about transition durations % are not reported exactly 
(see Section \ref{sec:node_attributes}).
In terms of the associated MDP, the learner solves partially-observable MDP (POMDP) \cite{kaelbling:aij98}, where partial observability concerns only the transition durations, and thus plays a minor role.

In the approach presented here, the discrete graph is used both for building the continuous graph and as a simulator of the learned strategy. Having a satellite simulator capable of getting the attitudes and transition times on-the-fly would remove the discrete graph building requirement, without significant change in our approach. At the time of writing, the discrete graph is precomputed by calling a closed-source proprietary satellite simulator, and there is no simple legal way to switch to the \textquote{on-the-fly} approach.
%As stated above, in this work we rely  on the discrete graph. We use it both for building the continuous graph and as a simulator of the learned strategy. Pre-building this discrete graph could be avoided by directly setting up the continuous graph from the problem data, and using a satellite simulator to get the attitudes and transition times on-the-fly, without significant change in our approach. At the time of writing, the discrete graph is precomputed by calling a closed-source proprietary satellite simulator, and there is no simple legal way to switch to the \textquote{on-the-fly} approach. 

%
 %because transition durations are not reported exactly (see Section \textquote{Node Attributes}).  

\section{Solution}
\label{sec:sol_technique}

%\begin{figure}[!htbp]
\begin{figure}[t]
  \centering
  \begin{tikzpicture}[scale=.6,every text node part/.style={align=center},every node/.style={scale=0.6}]
  	\tikzstyle{elem} = [draw=none,fill=none,text width=1.8cm]
  	
  	\begin{scope}
  		\node[draw,rounded corners, rectangle, minimum width=10cm, minimum height=3cm,fill=black!5]  (agent) at (4,0) {};
  		
  		\node[elem] (graphRewirer) at (0,0) {Graph\\Rewirer};
  		\node[elem] (nodeEmbed) at (3,.75) {Node\\Embedder};
  		\node[elem] (edgeEmbed) at (3,-.75) {Edge\\Embedder};
  		\node[elem,text width=1.2cm] (gnn) at (5.5,0) {GNN};
  		\node[elem] (valEstim) at (8,.75) {Value\\Estimator};
  		\node[elem] (actSelector) at (8,-.75) {Action\\Selector};
  		
  		%\node at (0,-.95) {\includegraphics[width=1.3cm]{fig/rewired.png}};
  		
  		\draw[->] (graphRewirer) -- (nodeEmbed);
  		\draw[->] (graphRewirer) -- (edgeEmbed);
  		\draw[->] (nodeEmbed) -- (gnn);
  		\draw[->] (edgeEmbed) -- (gnn);
  		\draw[->] (gnn) -- (valEstim) node [pos=.3, above] {\footnotesize graph\\ \footnotesize logit};
  		\draw[->] (gnn) -- (actSelector) node [pos=.3, below] {\footnotesize nodes\\ \footnotesize logits};;

  		\node at (-0.4,1.1) {Agent};
  		\draw[dashed] (-1,0.75) -- (.2,0.75);
  		\draw[dashed] (.2,0.75) -- (.2,1.5);
  	\end{scope}
  	
  	\begin{scope}[yshift=-5cm,xshift=2.5cm]
  		\node[draw,rounded corners, rectangle, minimum width=6cm, minimum height=2cm, fill=black!5] (simu)  at (1.5,0.5) {};

  		\node[elem,text width=6cm] (graphUpdate) at (1.5,0.2) {Discrete and Continuous Graphs Update};
  		%\node[elem,text width=2.2cm] (sysSimu) at (3,0) {System Simulator};
  		
  		%\begin{scope}[xshift=-.1cm]
  		%	\draw[red] (-.3,-1) circle (3pt);
  		%	\draw[red] (-.7,-.6) circle (3pt);
  		%	\draw[ForestGreen] (-.3,-.6) circle (3pt);
  		%	\draw[RoyalBlue] (-.7,-1) circle (3pt);
  		%	\draw (-.61,-1) -- (-.39,-1);
		%	\draw (-.61,-.6) -- (-.39,-.6);
  		%	\draw[dashed,red] (-.63,-.67) -- (-.37,-.93);
  		%\end{scope}
  	
  		%\node at (0,-.8) {\footnotesize $\Rightarrow$};

		%\begin{scope}[xshift=1.1cm]
		%	\draw[red] (-.3,-1) circle (3pt);
		%	\draw[red] (-.7,-.6) circle (3pt);
		%	\draw[thick,->,red] (-.63,-.67) -- (-.37,-.93);
		%	\draw[ForestGreen] (-.3,-.6) circle (3pt);
		%	\draw[RoyalBlue] (-.7,-1) circle (3pt);
		%	\draw (-.61,-1) -- (-.39,-1);
		%	\draw (-.61,-.6) -- (-.39,-.6);
		%\end{scope}

  		\node at (-0.6,1.1) {Simulator};
  		\draw[dashed] (-1.5,0.8) -- (.3,0.8);
  		\draw[dashed] (.3,0.8) -- (.3,1.5);
  	\end{scope}
  	
  	\draw (8.6,-.75) -- (9.5,-.75) -- (9.5,-4.5) node [midway, right] {\\ \\ Action\\(next\\acquisition)};
  	\draw[->] (9.5,-4.5) -- (simu.east);
  	
  	\draw (simu.west) -- (-1.5,-4.5) -- (-1.5,0) node [pos=.7, left] {\\ \\ \\ \\ \\ \\ Partial\\schedule\\(continuous\\graph)};
  	\draw[->] (-1.5,0) -- (agent.west);
  	
  	%\node[anchor=east] at (-1.5,-3) {\includegraphics[width=1.5cm]{fig/partial_schedule_graph.png}};
  	
  	\node[text centered, rounded corners, dotted, rectangle, draw, text width=1cm, minimum height=.75cm, fill=black!5] (ppo) at (4,-2.5)  {PPO};
  	
  	\draw[->,dotted] (9.5,-2.5) -- (ppo.east);
  	\draw[->,dotted] (-1.5,-2.5) -- (ppo.west);
  	
  	\draw[->,dotted] (ppo) -- (edgeEmbed);
  	\draw[->,dotted] (ppo) -- (nodeEmbed) node[pos=.15] {\footnotesize update parameters};
  	\draw[->,dotted] (ppo) -- (gnn);
  	\draw[->,dotted] (simu) -- (ppo) node[pos=.8,right] {\footnotesize reward};
  	
  	\draw[dotted] (8.6,.75) -- (10,.75) -- (10,-2.3);
  	\draw[->,dotted] (10,-2.3) -- (4.625, -2.3) node[pos=.8,above] {\footnotesize value};
  	
  \end{tikzpicture}
  \caption{Wheatley general architecture}
  \label{fig:archi}
\end{figure}

Following  \cite{wheatley-jssp}, we use a reinforcement learning setup where the agent receives continuous graphs representing partial schedules as input, selects the next observations to schedule, and updates its parameters based on the reward representing the cumulative utility of the schedules it produces.  For a given problem, a simulator is in charge of managing the different graphs and feeding the learner with the appropriate data.  The learner implements a policy, that is, a stochastic mapping from states $s$ to actions $a$ as defined above.  It learns a policy that maximizes the reward function over a base of real-world problems used as training set.  The policy has to be able to generalize to test problems, that is, exhibit good performances without further learning on a set of instances not seen before.

An overview of the architecture is shown in Fig.\ \ref{fig:archi}.  The graphs provided as input are processed by several elements.  First, the graph is transformed in order to allow bidirectional message-passing by the GNN, then simple networks produce node and edge embeddings.  Next the graph is processed through a Message-Passing Graph Neural Network (MP-GNN) to extract features capturing relevant information, and produce action probabilities. Finally, the RL algorithm updates the parameters of the whole system, embedders and GNN, based on the rewards received.  For ease of presentation, we first discuss the RL algorithm, then the embedders and GNN.  %These different elements are described in this section.  

\subsection{Reinforcement Learning}
\label{sec:reinforcement_learning}

% \begin{algorithm}
% \footnotesize{
% \caption{General algorithm\label{algo:general_algo}}
% %\DontPrintSemicolon 
%     Generate validation instances, compute heuristics, and measure OR-Tools performance on these instances \;
%     %\tcp{actor is $\thicksim$ current policy $\pi_{\theta}$}
%     Initialize the actor randomly\;
%     %\tcp{critic is $\thicksim$ value function estimator}
%     Initialize the critic randomly\;
%     \For{$i = 1, 2, \ldots N$}{
%         \tcp{Collect dataset}
% 	Generate train instances  \;
%         Collect trials data $\mathcal{D}_i = ((s_t, a_t, s_{t+1}, s_{t+1}), ...)$ using current actor to select actions \;
%         Compute returns on trials \;
% 	Compute advantages on trials using current critic \;
% 	Rewire graphs in trial data (???)\;
% 	\tcp{PPO update algorithm}
% 	\Repeat{max number of iterations or too large KL-divergence between current and updated policy} {
% 		Sample a minibatch of $n$ data points over shuffled collected data \;
% 		Update actor over the minibatch data towards advantage maximization \;
% 		Update critic by MSE regression \;
% 	}
% 	Evaluate current policy (actor) on validation instances \;
%     }
% }
% \end{algorithm}
	
%\paragraph{DRL Algorithm:}
As our core algorithm, we use the Proximal Policy Optimization (PPO) algorithm \cite{PPO} with action masking \cite{ppo_mask}, due to its relative simplicity and its good results on many  different problems. 

A peculiar aspect of the EOSP instances we have to solve is that the utility of different acquisitions may vary by up to 8 degrees of magnitude.  In fact, acquisitions are  grouped in 7 priority classes with utility value ranging from 1 to $10^8$.  The utility of the observations within a class of priority is equal to the value of that class, plus a small term depending on the predicted cloud coverage at the location of the acquisition (in order to favor acquisitions that are likely to happen with a clear sky). This generates instability in DRL algorithms (and in MDPs in general), as the low priority observations provide a reward that might be difficult to distinguish from  noise in the algorithm.  In addition, the critic must learn very large values, starting from very low values at initialization, and following tiny gradient steps.  This makes learning slow and inefficient.

We tried several approaches to handle this, including using a logarithmic scale and 2-hot encoding \cite{dreamerv3}.  In our current implementation, we simply divide each individual reward by the average utility of all candidate observations in the problem.  This is a simple way to remedy the issue of having to learn very large values,  but it does not fix the problem of the discrepancy between rewards (unless some extreme priority classes are not represented in the problem instance).  We are currently examining optimization with lexicographic preferences \cite{lexicoRL}.  
%We are currently exploring models of learning with lexicographic preferences
%\todo[inline,linecolor=blue,backgroundcolor=blue!25,bordercolor=blue]{Parler des utilités  1 120 3300 6600 1e4 1e6 1e8 , avec des grandes variances du nombre d'acq par classe suivant les problèmes. Design de ces utilités car greedy en dessous => on veut quasiment forcer un ordre global sur les insertions
%1 -> reward unitaire
%2 -> méthode Antoine

\subsection{GNN Implementation}

\paragraph{Node attributes}  
\label{sec:node_attributes}

To inform the learner, we label each node $i$ of the continuous graph $\mathcal{G}^C$ with the visibility window $W_i$. The continuous graph does not bring any information about the transition duration to the learner. To compensate for this, each node $i$ of the continuous graph is labeled with information about the satellite attitude while performing observation $i$, namely, the min, max and average pitch and roll angles of the satellite over the observation VTW. Although this information is not sufficient to recover the exact duration of transitions, it allows the learner to infer them closely enough to perform well, as shown in our simulation results.

\paragraph{Graph rewiring}  

A Message-Passing Graph Neural Network (MP-GNN) \cite{gin} uses a graph structure as a computational lattice.  It propagates information, represented as messages, along the oriented graph edges only.  In our case, if an MP-GNN uses only the EOSP continuous graph edges, then information cannot flow from future acquisitions to the present choice of the next acquisition.  This is definitely not what is desired: the agent should choose the next observation to schedule based on its effect on future conflicts.  In other words, we want information to go from future to present tasks.  Therefore, we edit the input graph before it can be used by the MP-GNN. This is known in the MP-GNN literature as \textquote{graph rewiring}.  

For every (precedence) edge in the continuous graph, a link pointing in the other direction is added to the rewired graph (reverse-precedence).  Different edge types are defined for precedence and reverse-precedence edges, to enable the learned operators to differentiate between chronological and reverse-chronological links.  The system learns to pass information in a forward and backward way, depending on what is found useful during learning.  %In addition, self-loops are added to each node, to avoid a local node label embedding being too diluted by the messages received from other nodes.

\paragraph{Embeddings}
  
A graph embedder builds the rewired graph by adding edges.  It embeds node attributes (VTW, attitude stats) using a learnable MLP, and edge attributes (type of edge) using a learnable embedding. The output dimension of embeddings is an open hyper-parameter \emph{hidden\_dim}.  We found a size of 64 being good in our experiments.

\paragraph{Graph pooling}

A node is added and connected to every other node to allow collecting global information about the entire graph, as opposed to the local information associated with the nodes of the original graph.  It is used by the critic to estimate the value of the graph as a whole.  It is also used by the actor, where the global graph encoding is concatenated to each node embedding.  Indeed, messages are passed by the MP-GNN algorithm only between immediate neighbors.  Therefore, a network of depth $n\_layers$ is able to anticipate only $n\_layers$ observations ahead.  Having the global node embedding concatenated to each node embedding compensates for this, allowing the current decision to take into account the entire graph.\footnote{Adding such a kind of node to the graph is equivalent to learning a custom graph pooling operator.}

\paragraph{GNN}  

As a message passing GNN, we use EGATConv from the DGL library \cite{dgl}, which enriches GATv2 graph convolutions \cite{gatv2} with edge attributes.  We used 4 attention heads, leading to an output of size $4 \times$\emph{hidden\_dim}.  This dimension is reduced to \emph{hidden\_dim} using learnable MLPs, before being passed to the next layer (in the spirit of feed-forward networks used in transformers \cite{attention}). The output of a layer can be summed with its input using residual connections \cite{residual}. For most of our experiments, we used 10 such layers.  The message-passing GNN yields a value for every node and a global value for the graph (from the graph pooling node).

\paragraph{Action selection}

Action selection aims at computing action probabilities given the node values (logits) output by the GNN.  We can either use the logits output from the last layer of the GNN, or use a concatenation of the logits output from every layer.  We chose to concatenate the global graph logits of every layer, leading to a data size of $((n\_layers + 1) \times $\emph{hidden\_dim}$) \times 2$ per node, where \emph{hidden\_dim} is the dimension of the embeddings.  This dimension is reduced to 1 using a learnable linear combination, that is, a minimal case of a Multi-Layer Perceptron (MLP).   We did not find using a larger MLP to be useful.  Finally,  a distribution is built upon these logits by normalizing them, and using action masks to remove actions that are not feasible in the current state.  As node numbers correspond to action/acquisition numbers, we directly have the action identifier when drawing a value from the distribution.

\paragraph{Dealing with different problem sizes}

The GNN outputs a logit per node, and there is a one-to-one mapping between nodes and actions whatever the number of nodes/actions.  Learning the best action boils down to node regression, with target values being given by the reinforcement learning loop.  Internally, the message passing scheme collects messages from all neighbors, making the whole pipeline agnostic to the number of nodes. 

\paragraph{Connecting to PPO}

In most generic PPO implementations, the actor (policy) consists of a feature extractor whose structure depends on the data type of the observation, followed by a MLP whose output dimension matches the number of actions.  The same holds for the critic (value estimator), with the difference that the output dimension is 1.  Some layers can be shared (the feature extractor and first layers of the MLPs).  In our case, we do not want to use such a generic structure because we have a one-to-one matching from the number of nodes to the actions.  We thus always keep the number of nodes as a dimension of the data tensors.

\subsection{Inference Time Search}
Ideally, the agent should be able to select the best action simply by selecting the argmax of the scores of the candidate actions. But several reasons may lead to suboptimal behavior doing so: the learning may not be finished (for instance if a plateau is reached during learning phase), the GNN may not be able to retain information far enough into the future of the current schedule (due to GNN over-smoothing or over-squashing, and due to the fact that we choose actions chronologically), or the agent may not have enough generalization ability (it learns on a given set of scenarios, and it is difficult to measure the closeness of these setups to evaluation/real-world setups). For all these reasons, we propose to perform search at inference time, meaning after the learning phase of the agent, in order to further refine the policy. Several well-known techniques are possible, like beam search or Monte-Carlo Tree Search. We found beam search to be hard to evaluate because of the very large branching factor we face, leading to exploring very few beams in reasonable time/memory. We thus conducted some experiments with MCTS, using PUCT bound \citep{Silver_2016}.  

For doing so, we allocate some budget for MCTS exploration. One key aspect of MCTS is the way to set the initial value of expanded nodes; in most MCTS uses, random rollouts are performed until the end in order to estimate a so-called empirical mean. In our case, as we have already learned a value function for the states in the learning phase, we can directly plug this value as an initial estimate. 
Estimating the Q-value during selection also requires using a value for unexplored children nodes. Traditional MCTS uses an infinite value that forces in-breadth exploration of children. This is detrimental when the simulation budget is low wrt. the action space, as is in EOSP tasks. Here we copy the parent's estimated value to the unexplored children \citep{DBLP:journals/corr/abs-2104-04278}.

%We present some results of these experiments in the following section. 

\section{Experiments}
\label{sec:expes}

%\subsection{Data and baselines}

We use a set of real-life non-public problems consisting of 100 to 1500 candidate acquisitions to be scheduled over a single orbit. From those, there's room for only about 50 of them to be selected for execution, yielding a largely over-subscribed problem for every orbit.  As explained in Section \ref{sec:problem_representation}, problems are given in the form of discrete graphs. The simulator uses this graph to compute and maintain the continuous-time graph. %\footnote{It is worthwhile to underline that a much more efficient way would be to use only the continuous graph in the simulator, using a spatial mechanics simulator to compute on-the-fly transition times. This  approach was not possible due to IP issues on the spatial mechanics simulator.} %In other words, the discrete graph is an approximation (due to discretization) of all possible courses of actions.
To provide intuition on the difficulty of the problem, Table \ref{tab:problem_sizes} shows some statistics on a few test problems and their representation as graphs.

\begin{table*}[htbp!]
  \centering
  \small
  \setlength{\tabcolsep}{5pt}
  \begin{tabular}{ccccccc}
    & \multicolumn{3}{c}{\# Nodes} & \multicolumn{3}{c}{\# Edges} \\
    \cmidrule(r){2-4} \cmidrule(r){5-7}
    \# Acquisitions & Discrete graph & Continuous graph & Ratio & Discrete graph & Continuous graph & Ratio\\
    \midrule
    106 & 10297 & 106 & 97 & 835566 & 9273 & 90\\
    308 & 52020 & 308 & 169 & 12598738 & 81244 & 155 \\
    508 & 46589 & 508 & 92 & 14842035 & 225398 & 66 \\
    809 & 59583 & 809 & 74 & 28015753 & 447945 & 63 \\ 
   1074 & 94071 & 1074 & 88 & 58343397 & 741634 & 79\\
    \bottomrule
  \end{tabular}
  \caption{Representative problem sizes}
  \label{tab:problem_sizes}
\end{table*}

We compare our DRL approach, Wheatley, to two solutions currently being used for operating such satellites.

\paragraph{Greedy algorithm} It is the algorithm currently used for real-world operations. It greedily selects acquisitions to add to the schedule based on their utility, and inserts them in the plan if possible. Previously selected tasks may be slightly postponed, but never canceled. % See Alg.\ \ref{algo:greedy} adn \ref{algo:check-chainability}.  Note that it does not require the discrete graph.

\paragraph{RAMP} \cite{RAMP}: It is an implementation of a Dijkstra search algorithm in the discrete graph.  Although based on an admissible algorithm (Dijkstra), RAMP is not guaranteed to find the absolute optimal  schedule.  This is due to the exclusion links between nodes of the discrete graph that represent different start dates of the same task.  Nevertheless, RAMP constantly provides the best known schedules on real problems.  Unfortunately, its complexity prevents using it for real-time operations.  Therefore, it is used as a reference in these simulation results.
%\todo[inline,linecolor=blue,backgroundcolor=blue!25,bordercolor=blue]{Vincent et Jonathan : vérifier ce qui est diffusable ou pas sur le greedy, expliquer RAMP}

\subsection{Unitary Score}

\begin{figure}[htb]
  \centering
	\includegraphics[width=8cm]{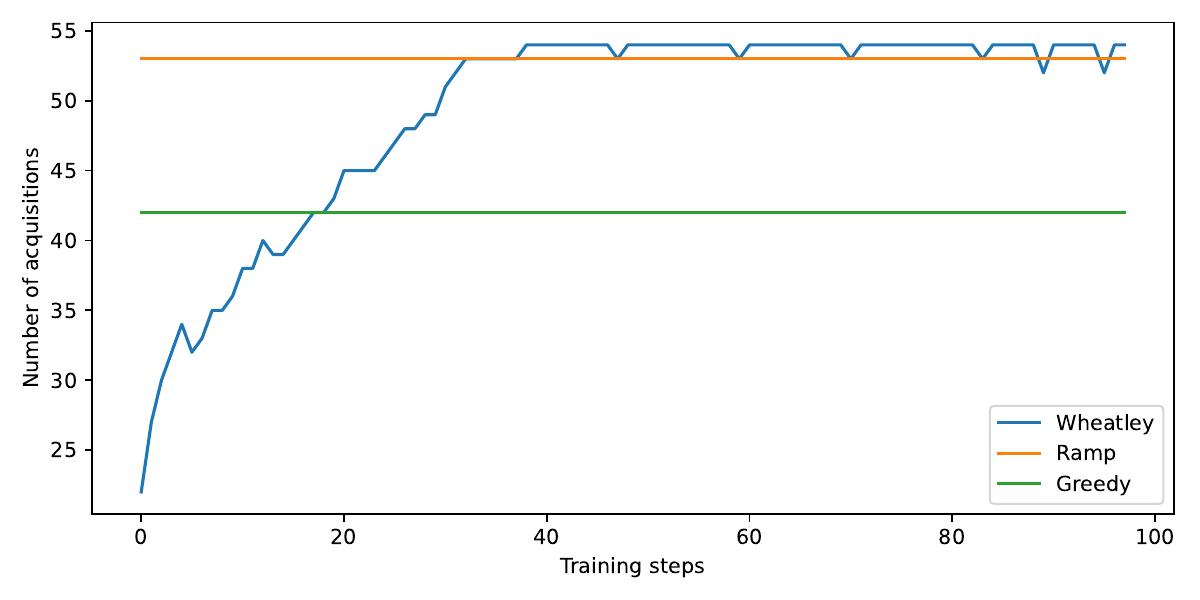}
	\caption{Unitary scores:  single problem of 106 acquisitions. }
	\label{fig:unitary_scores_single}
\end{figure}

First, we compare our approach to baselines on a relaxed problem: we try to maximize the number of acquisitions scheduled, irrespectively of their priority or utility.  This measure of performance is insensitive to the large gaps in acquisitions utility discussed in Section \ref{sec:reinforcement_learning}.  We run two  experiments:

\paragraph{Single problem}
First, we want to measure if our models and algorithms can possibly achieve competitive performance on a given problem.  We train our learner on a single problem with a total of 106 acquisitions and let it overfit as much as needed, as long as it achieves great performance.   As illustrated in Fig.~\ref{fig:unitary_scores_single}, we observe that it is indeed able to outperform both the greedy algorithm and RAMP scores.  This result shows that our architecture is able to implement very powerful policies.  In the next set of experiments, we put it to the challenge of a realistic learning environment.

%\begin{figure}
%	\includegraphics[width=\textwidth]{fig/unitary_single_scores.pdf}
%	\caption{Training on a single problem of 106 acquisitions, using unitary score.}
%	\label{fig:unitary_single_scores}
%\end{figure}

\begin{figure}[htb]
\centering
\includegraphics[width=8cm]{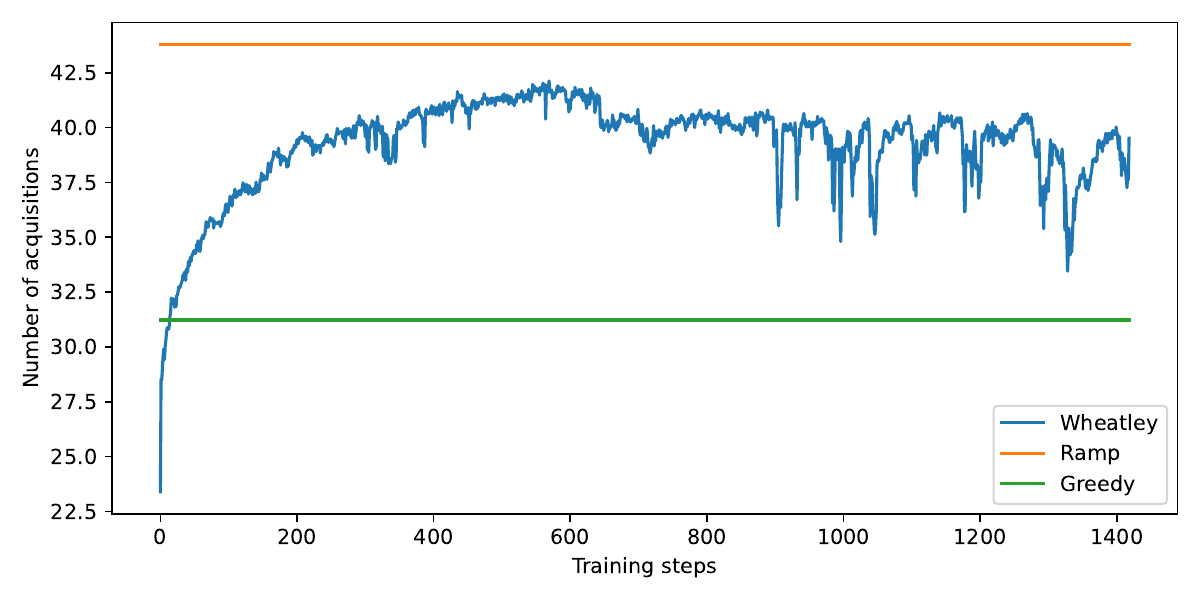}
\caption{Unitary scores: performance on 31 unseen problems (training on 128 different problems)}
\label{fig:unitary_scores_gene}
\end{figure}

\begin{figure}[htb]
   \centering
	\includegraphics[width=8cm]{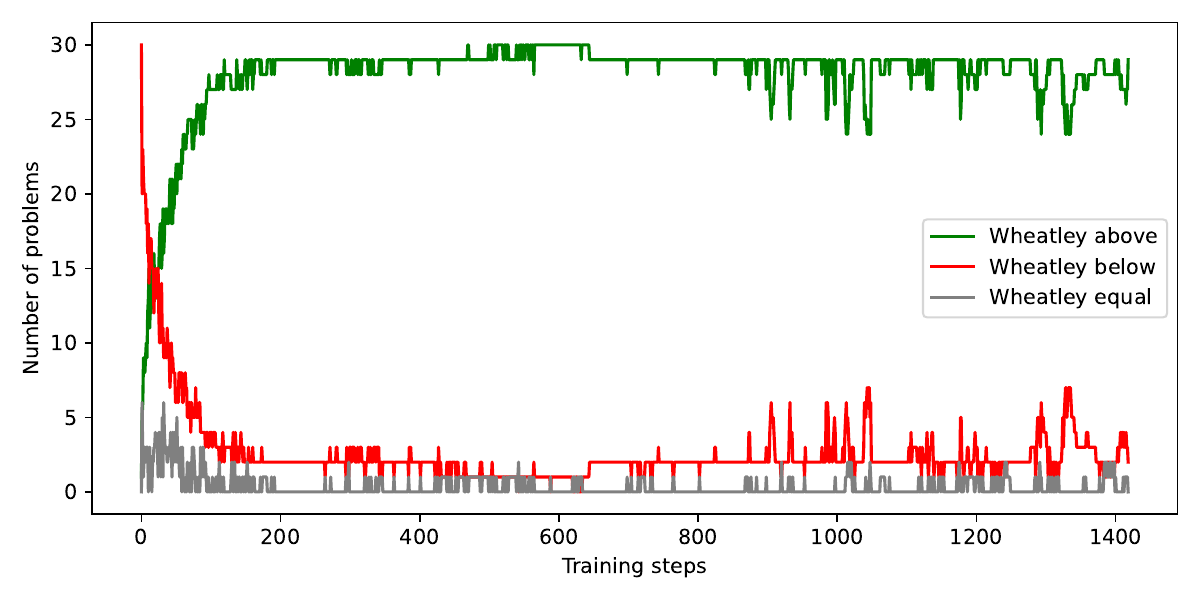}
	\caption{Unitary scores: Wheatley vs. Greedy}
	\label{fig:unitary_generalization_greedy}
\end{figure}

\begin{figure}[htb]
   \centering
	\includegraphics[width=8cm]{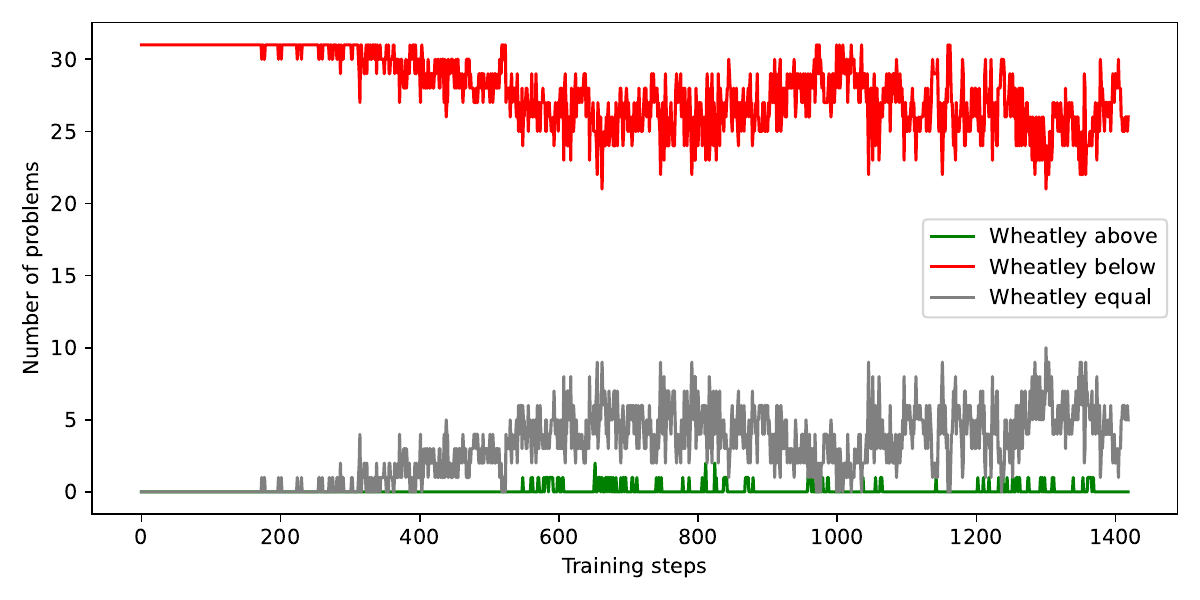}
	\caption{Unitary scores for Wheatley vs. Ramp}
	\label{fig:unitary_generalization_ramp}
\end{figure}

\paragraph{Generalization}
To measure the ability to transfer knowledge from one task to another, we train on 128 problems of about 100 acquisitions and test on 31 unseen problems of similar size.  The learning curve of Fig.~\ref{fig:unitary_scores_gene} shows the evolution of the performance on the test set, as learning progresses. It peaks at around 600 training steps before slowly decreasing due to overfitting.  We also measure the number of times where Wheatley's performances are above, below or equal to the greedy algorithm (Fig. \ \ref{fig:unitary_generalization_greedy}) and  RAMP (Fig.\ \ref{fig:unitary_generalization_ramp}) on the 31 test problems.  This shows that our system is able to generalize to unseen problems, outperforming the currently deployed solution.  %Finally, Fig.\ \ref{fig:trajectory} shows examples of solutions produced for Greedy and Wheatley.   We can see that Weathley finds a smoother and more efficient trajectory for the satellite.

%\begin{figure}
%	\includegraphics[width=\textwidth]{fig/unitary_generalization_scores.pdf}
%	\caption{Training on 128 problems and testing on 31 unseen problems, trying to maximize unitary score.}
%	\label{fig:unitary_generalization_scores}
%\end{figure}

%\begin{figure}
%	\includegraphics[width=\textwidth]{fig/unitary_generalization_greedy.pdf}
%	\caption{Comparison of Wheatley and Greedy unitary scores on 31 test problems.}
%	\label{fig:unitary_generalization_greedy}
%\end{figure}
%\begin{figure}
%	\includegraphics[width=\textwidth]{fig/unitary_generalization_ramp.pdf}
%	\caption{Comparison of Wheatley and Ramp unitary scores on 31 test problems.}
%	\label{fig:unitary_generalization_ramp}
%\end{figure}

%\begin{figure}
%	\begin{center}
%	\includegraphics[width=6cm]{fig/trajectory.pdf}
%	\end{center}
%	\caption{Example of trajectory found by Wheatley on a 87 acquisitions problem compared to the Greedy trajectory.}
%	\label{fig:trajectory}
%\end{figure}

\subsection{General Utility}

In our second set of experiments, we take into account the utility of observations and aim at maximizing the cumulative utility of all the observations included in the final schedule, as in the full-fledged MDP framework presented in Section \ref{sec:reinforcement_learning}.  As before, we perform two sets of experiments: one where the learner is free to overfit on a single problem to reach its best performance, and one aiming at measuring its ability to generalize. 

\begin{figure}[htb]
  \centering
	\includegraphics[width=8cm]{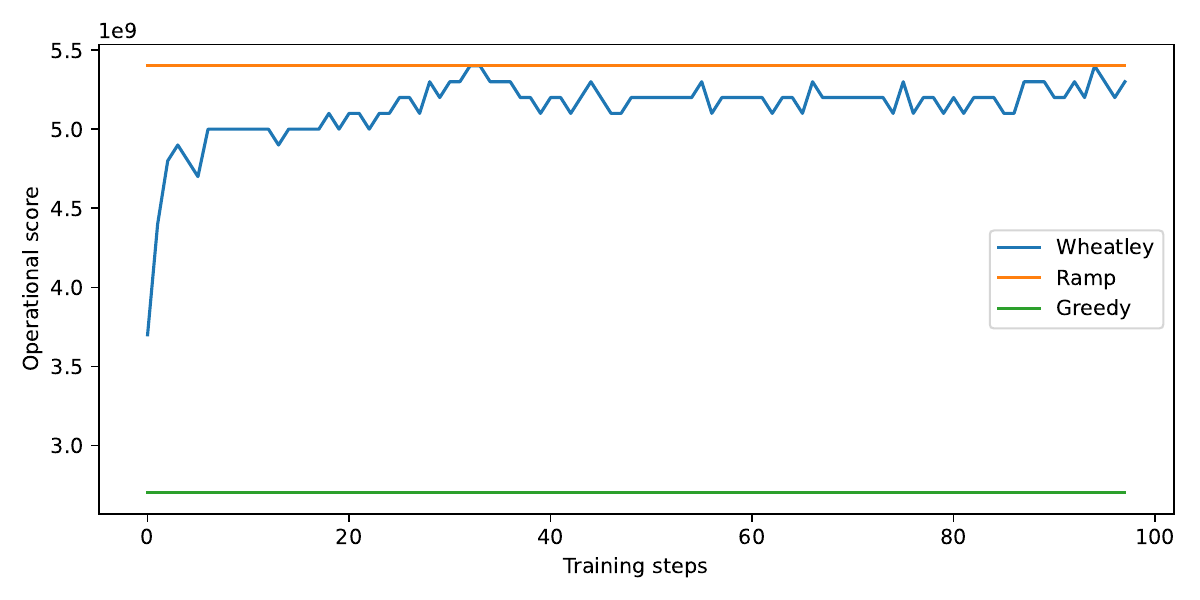}
	\caption{Utilities obtained when training on a single problem with 88 acquisitions}
	\label{fig:operational_scores_single}
\end{figure}

\paragraph{Single Problem}
Our test on a single problem with a total of 88 candidate acquisitions shows that our system is able to outperform the greedy algorithm and reach the score of RAMP (Figure~\ref{fig:operational_scores_single}).  This proves the suitability of the architecture for the full MDP set up.

%\begin{figure}
%	\includegraphics[width=\textwidth]{fig/operational_single_scores.pdf}
%	\caption{Training on a single problem of 88 acquisitions, trying to maximize operational score.}
%	\label{fig:operational_single_scores}
%\end{figure}

\begin{figure}[htb]
  \centering
	\includegraphics[width=8cm]{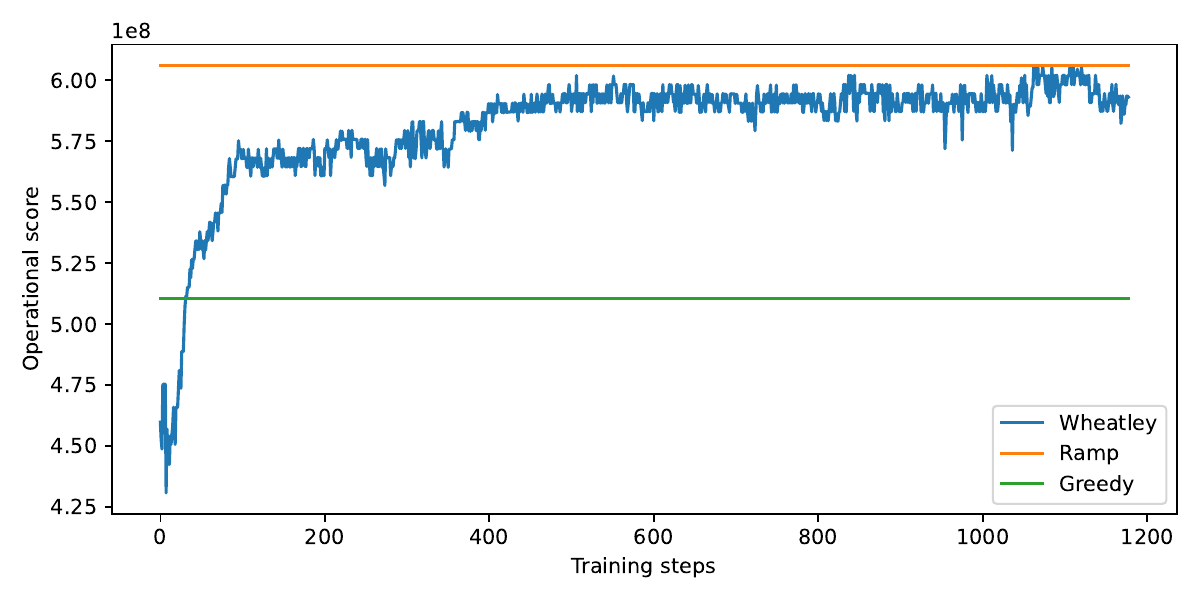}
	\caption{Utilities obtained averaging on 27 unseen problems after a training on 639 different problems}
	\label{fig:operational_scores_gene}
\end{figure}

\begin{figure}[tbh]
  \centering
	\includegraphics[width=8cm]{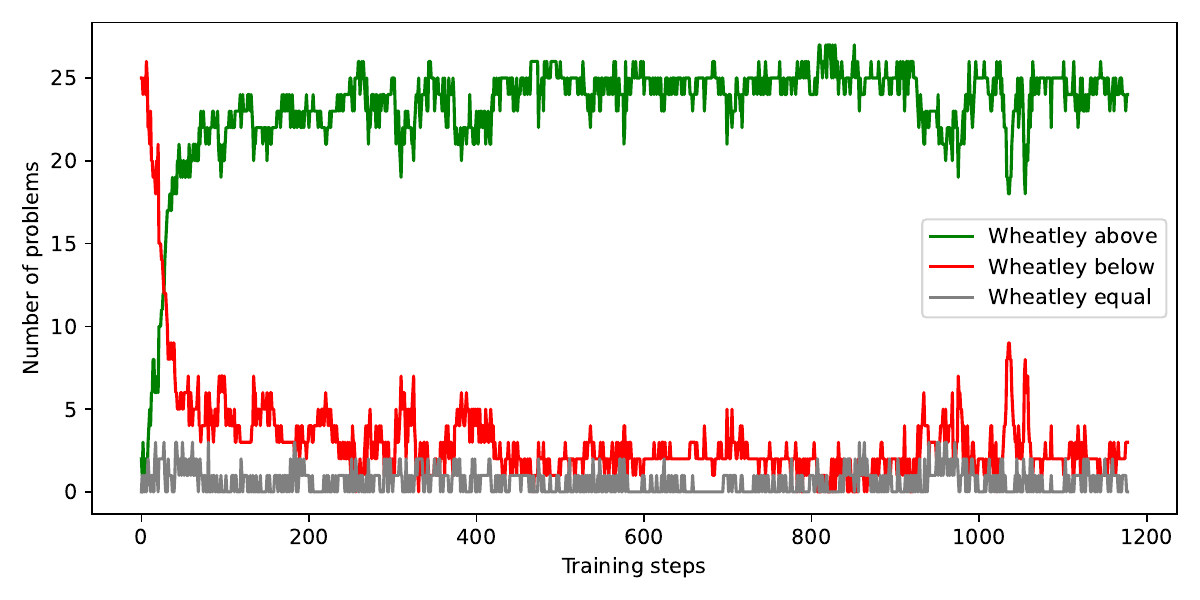}
	\caption{Utilities for Wheatley vs. Greedy}
	\label{fig:operational_generalization_greedy}
\end{figure}

\begin{figure}[tbh]
  \centering
	\includegraphics[width=8cm]{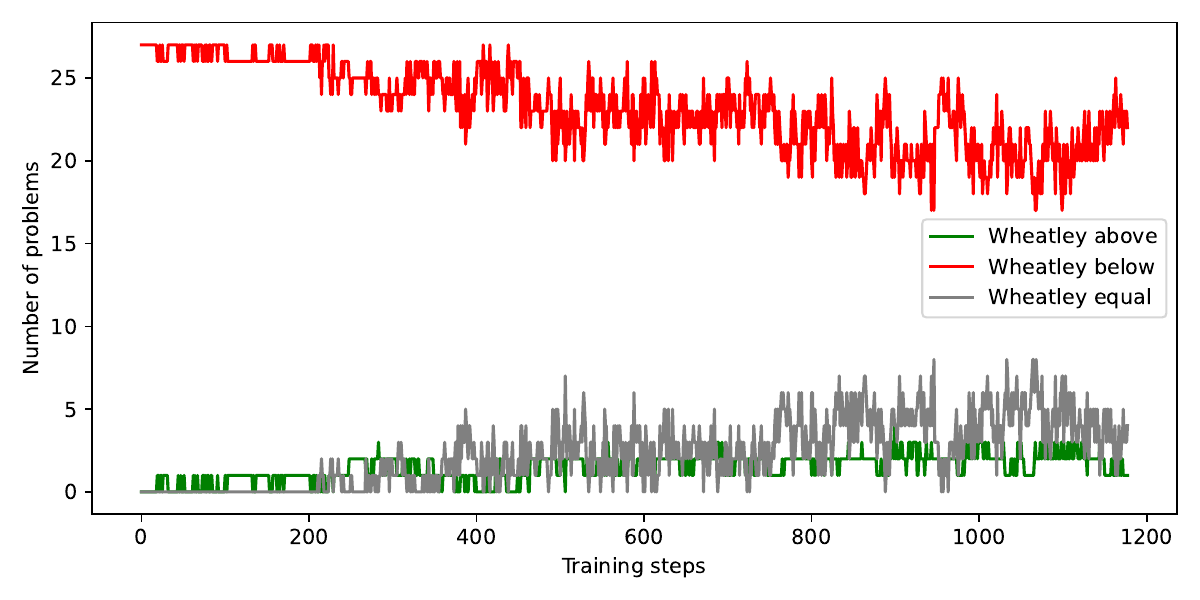}
	\caption{Utilities for Wheatley vs.  Ramp}
	\label{fig:operational_generalization_ramp}
\end{figure}

\begin{figure}[tbh]
	\begin{center}
	\includegraphics[width=8cm]{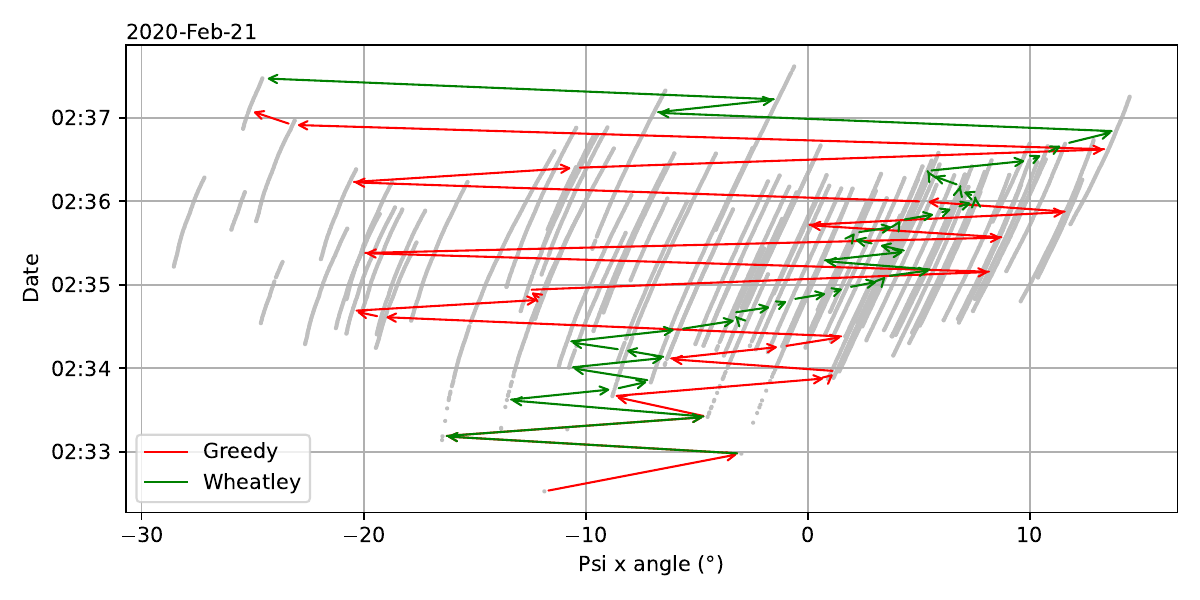}
	\end{center}
	\caption{Trajectories found by Wheatley and Greedy approaches on a 87 acquisitions instance.}
	\label{fig:trajectory}
\end{figure}

\paragraph{Generalization}
We train on 639 problems of about 100 acquisitions and test on 27 unseen problems of similar size.  The learning curves are displayed in Figure \ref{fig:operational_scores_gene} and show that the learner is able to generalize.  The plot showing the number of times where Wheatley is above, below or equal to the greedy solution are presented in Fig.\ \ref{fig:operational_generalization_greedy} and same for RAMP are presented in Fig.\ \ref{fig:operational_generalization_ramp}.   We see that Wheatley outperforms the deployed solution and approaches the best known performances, in a realistic set-up where problems are not known in advance.
Fig.\ \ref{fig:trajectory} shows examples of satellite trajectories produced by Greedy and Wheatley. It represents the roll angle (X-axis) at each acquisition date (Y-axis). Acquisition requests are depicted in gray during their VTW. Manoeuvers between acqusitions are the colored arrows describing the satellite depointing angle along time. We can see that Wheatley finds a smoother and more efficient sequence for the satellite.
%
%Finally, in order to evaluate generalization abilities, we also evaluate the learned agent on problems of different sizes. Results can be seen on table \ref{tab:generalization-scores}.

Table \ref{tab:generalization-scores} shows comprehensive results for the agent trained on problems of size 100, evaluated on different sizes of problems. The last line is an evaluation on instances with many conflicts, where RAMP performs worse than the greedy algorithm. Results show that Wheatley performs very well on not too large instances but is outperformed by the greedy approach on the largest instance. However, it is quite competitive in the case of highly conflictual instances, which is promising for future works.

%\begin{figure}
%	\includegraphics[width=\textwidth]{fig/operational_generalization_scores.pdf}
%	\caption{Training on 639 problems and testing on 27 unseen problems, trying to maximize operational score.}
%	\label{fig:operational_generalization_scores}	
%\ref{fig:operational_generalization_ramp} on the test problems.
%\end{figure}

%\begin{figure}
%	\includegraphics[width=\textwidth]{fig/operational_generalization_greedy.pdf}
%	\caption{Comparison of Wheatley and Ramp operational scores on 27 test problems.}
%	\label{fig:operational_generalization_greedy}
%\end{figure}
%\begin{figure}
%	\includegraphics[width=\textwidth]{fig/operational_generalization_ramp.pdf}
%	\caption{Comparison of Wheatley and Ramp operational scores on 27 test problems.}
%	\label{fig:operational_generalization_ramp}
%\end{figure}


\begin{table*}[tb!]
  \centering
  \small
  \setlength{\tabcolsep}{5pt}
  \begin{tabular}{cccccccc}
    \multicolumn{2}{c}{Instances Set} & \multicolumn{3}{c}{Average Utility Score} & \multicolumn{2}{c}{Avg. Scores Ratios}\\
    \cmidrule(lr){1-2} \cmidrule(lr){3-5} \cmidrule(lr){6-7}
     \#Acq. & \#Instances &  Wheatley & Greedy &  Ramp  &  $\frac{\text{Wheatley}}{\text{Greedy}}$ &  $\frac{\text{Wheatley}}{\text{Ramp}}$ \\
    \midrule
    100 & 27 & 605,732,913 & 510,488,439 & \textbf{605,894,711}  & \textbf{1.1788} & 0.9901\\
    300 & 30  & 239,934,939 & 202,565,994 & \textbf{263,514,132} & \textbf{1.2560} & 0.9420\\
    508 & 1 & 221,226 & 142,487 & \textbf{308,355}  & \textbf{1.5526} & 0.7174\\
    809 & 1 & 52,000,039 & 49,000,279 & \textbf{59,000,286}  & \textbf{1.0612} & 0.8814\\
    1074 & 1  & 2,910,000,156 & 2,225,103,224 & \textbf{4,124,116,197} & \textbf{1.3078} & 0.7056\\
    1591 & 1  & 220,734 & 307,159 & \textbf{393,214}& 0.7186 & 0.5614\\
    \midrule
    100 & 10  & \textbf{689,578,101} & 688,941,262 & 678,921,586 & 0.9996 & \textbf{1.0115}\\
    \bottomrule
  \end{tabular}
  \caption{Average scores obtained when generalizing on different instances sets. }
  \label{tab:generalization-scores}
\end{table*}

\subsection{Inference Time MCTS}
Results of policy search after learning is shown on Table \ref{tab:mcts}, though on a different dataset\footnote{The satellite target was updated during the work with support for a larger set of priority levels.} than in Table \ref{tab:generalization-scores}. In this new dataset, the average number of selectable acquisitions is about 15, but the number of priority levels is higher, leading to larger operational scores when selecting high priorities. The displayed scores are the mean over the test set.

The budget is the number of trials done (\textit{i.e.} the number of paths leading to tree expansion and backtrack of the estimates along the path) before selecting an action. MCTS uses the learned value network to get initial evaluation of new nodes. Two criterions are of interest: the mean absolute score on the test set, and the number of times where the MCTS search  is better than  RAMP reference. The results show that using MCTS at resolution times consistently improves found solutions over \textquote{vanilla} Wheatley.

It is worth noting that such budgets are small compared to the problem sizes, as the problems exhibit a large branching factor. For instance, in a problem of 100 acquisitions, a budget of 100 allows only tree exploration until a depth of 6, which is very far from the maximal depth of the tree (which could be up to 100 if all acquisitions are selectable).  We are limited to such small budgets in order to perform search phase in sensible time: due to the update of discrete graphs and computations of continuous graphs at every node, this search phase duration is far from negligible, as shown in the results table. %Typically, a budget of 100 takes 138 minutes (using 8 threads) and a budget of 1000 takes 1248 minutes (using 4 threads). 

\begin{table}[tb!]
\centering
\small
\begin{tabular}{rccc}
 & score ($\times10^{17}$)& $\geq$ RAMP (\%)  & time (min) \\
 \hline
 RAMP &  8.62747 & 100 & \\
 Wheatley &  8.38172 & 28.7356& \\
 \hline
 MCTS(100) & 8.45111 & 37.9310 & 138\\
 MCTS(1000)  & 8.56566 & 43.6782 & 1248\\
 \end{tabular}
 \caption{MCTS(budget) results}
 \label{tab:mcts}
\end{table}

\section{Conclusion}
\label{sec:conclusion}

We showed that DL-based approaches to the EOSP challenge some of the best known techniques.  There are several perspectives we are currently exploring to extend this work.  First, as stated before, we are trying to take advantage of the large discrepancy in acquisition utility by using lexicographic RL algorithms such as \cite{lexicoRL}.  Scheduling tasks by decreasing priority would provide stronger guarantees to find the optimal schedule.  To achieve this, schedules must be built in a non-chronological order, which is not the case in our current implementation.   Currently, we choose the next acquisition to insert just after the last inserted one, using some foresight given by the GNN. This foresight is limited by the number of layers of the GNN.  As we said, the discrete-time graph is tailored  for state-space search and chronological insertion. Future work will consider developing an alternative continuous-time graph representation of the EOSP where observations can be added to the schedule in any order, using Simple Temporal Networks \cite{stn}.  Such work will open promising avenues for using lexicographic preferences.
% Therefore, we have developed an alternative continuous-time graph representation of the EOSP where observations can be added to the schedule in any order, using Simple Temporal Networks \cite{stn}.  This work opens promising avenues for using lexicographic preferences.  It will be presented in forthcoming publications.
%\paragraph{Perspectives}

%There are several avenues that we are currently exploring.  As stated before, 

%We would like to implement a real lexicographical reward, in the spirit of \cite{lexicoRL}.

%We are investigating non-chronological acquisition selection. In our current implementation, we choose acquisition to insert just after the last inserted one, using some foresight given by the GNN. This foresight is limited by the number of layers of the GNN.  Another option would be to be able to insert any acquisition,  which creates the need for being able to insert other acquisitions before previously selected ones. In order to do so, the continuous graph should be augmented with a Simple Temporal Network \cite{stn}, and acquisition insertion would lead to time-window constraint propagation. In our opinion, this would give more flexibility to the agent and alleviate GNN horizon bottleneck. 

%\bibliographystyle{plain}
\bibliography{Wheatley_eosp}

\end{document}